\documentclass{article}
\usepackage{ijcai18}
\usepackage{times}

\usepackage{stmaryrd} 
\usepackage{epsfig,graphicx} 
\usepackage{amsthm,amsmath,amssymb}
\usepackage{bbm}
\usepackage{latexsym}
\usepackage[norelsize,ruled,vlined,linesnumbered]{algorithm2e}
\usepackage{tikz}
\usepackage{mathrsfs}
\usepackage{hyperref}
\usetikzlibrary{patterns}
\usetikzlibrary{decorations.pathreplacing}

\long\def\comment#1{}

\newtheorem{definition}{Definition}
\newtheorem{theorem}{Theorem}

\newtheorem{lemma}{Lemma}

\newtheorem{example}{Example}
\newenvironment{proofsketch}{\noindent \textit{Proof}\textit{ (Sketch)}.}{$\Box$\\ }

\newenvironment{theorem-appendix}{\vspace*{0.25cm}\textit{Theorem} }{} 
\newenvironment{proposition-appendix}{\vspace*{0.25cm}\textit{Proposition} }{} 
\newenvironment{lemma-appendix}{\vspace*{0.25cm}\textit{Lemma} }{} 

\newcommand{\vect}{\mathbf}

\newcommand{\SC}{\textsc}
\newcommand{\SF}{\textsf}

\newcommand{\LA}{\langle}
\newcommand{\RA}{\rangle}
\newcommand{\ra}{\rightarrow}

\newcommand{\REST}{\hspace{-0.05in}\restriction}

\newcommand{\CONST}{\textsc{const}}
\newcommand{\TERMS}{\textsc{terms}}
\newcommand{\NULLS}{\textsc{nulls}}
\newcommand{\VAR}{\textsc{var}}
\newcommand{\UVAR}{\forall\mbox{-}\textsc{var}}
\newcommand{\EVAR}{\exists\mbox{-}\textsc{var}}
\newcommand{\MVAR}{\textsc{m-var}}
\newcommand{\ARG}{\textsc{arg}}

\newcommand{\tgd}{\sigma}
\newcommand{\tgds}{\Sigma}
\newcommand{\ATOMS}{\textsc{atoms}}

\newcommand{\REL}{\textsc{rel}}
\newcommand{\FCTGDS}{\textsc{fc-tgds}}
\newcommand{\TG}{\textsc{tg}}

\newcommand{\CYCNULL}{\textsc{cyc-null}}
\newcommand{\DOM}{\textsc{dom}}

\newcommand{\atoma}{\mathbf{a}}
\newcommand{\atomb}{\mathbf{b}}
\newcommand{\atomc}{\mathbf{c}}
\newcommand{\atomd}{\mathbf{d}}
\newcommand{\vatoma}{\overrightarrow{\mathbf{a}}}
\newcommand{\vatomb}{\overrightarrow{\mathbf{b}}}
\newcommand{\vatomc}{\overrightarrow{\mathbf{c}}}
\newcommand{\vatomd}{\overrightarrow{\mathbf{d}}}
\newcommand{\BD}{\textsc{bd}}
\newcommand{\NULLSET}{\textsc{nullset}}
\newcommand{\HD}{\textsc{hd}}

\newcommand{\LINK}{\textsc{link}}

\newcommand{\CHASE}{\textsf{chase}}

\newcommand{\LEVEL}{\textsc{level}}
\newcommand{\MAX}{\textsc{max}}

\title{A New Decidable Class of Tuple Generating Dependencies: 
       The Triangularly-Guarded Class}
\author{Vernon Asuncion and Yan Zhang\\
	    School of Computing, Engineering and Mathematics\\
	    Western Sydney University, Australia}

\begin{document}

\maketitle

\begin{abstract}
	In this paper we introduce a new class of tuple-generating dependencies (TGDs) 
	called {\em triangularly-guarded} TGDs, which are TGDs with certain restrictions on the atomic derivation track
	embedded in the underlying rule set. We show that conjunctive query answering under
	this new class of TGDs is decidable. We further show that this new class strictly contains
	some other decidable classes such as weak-acyclic, guarded, sticky and shy,
	which, to the best of our knowledge, provides a unified representation of all these aforementioned classes.		 
\end{abstract}

\section{Introduction}\label{intro}

In the classical \emph{database management systems} (DBMS) setting,
a query $Q$ is evaluated against a database $D$. However, it 
has come to the attention of the database community the
necessity to also include ontological reasoning and description
logics (DLs) along with standard database techniques
\cite{CalvaneseGLLR07}. As such,
the \emph{ontological database management systems} (ODBMS) has arised.  
In ODBMS, the classical 
database is enhanced with an ontology \cite{BaaderBL16} in the form
of logical assertions that generate new intensional
knowledge. A powerful form of such logical
assertions is the tuple-generating
dependencies (TGDs), i.e.,
Horn rules extended by allowing 
existential quantifiers to appear in the rule heads
\cite{Cabibbo98,Patel-SchneiderH07,CaliGL09}.
 
Queries are evaluated against a database $D$ and set of TGDs $\tgds$
\big(i.e., $D$ $\cup$ $\tgds$\big) rather than just $D$, as in the classical setting.
Since for a given database $D$, a set $\tgds$ of TGDs, and a conjunctive query $Q$, the problem of determining if 
$D$ $\cup$ $\tgds$ $\models$ $Q$, called the \emph{conjunctive query answering} (CQ-Ans) problem,  
is \emph{undecidable} in general \cite{BeeriV81,BagetLMS11,Rosati11,CaliGP12,CaliGK13}, a major research
effort has been put forth to identifying syntactic conditions on TGDs for which CQ-Ans is decidable.
Through these efforts, we get the decidable syntactic classes:
\emph{weakly-acyclic} (\SC{wa}) \cite{FaginKMP05}, 
\emph{acyclic graph of rule dependencies} (a\SC{grd}) \cite{BagetLMS11}, 
\emph{linear}, \emph{multi-linear}, \emph{guarded}, \emph{weakly-guarded} (\SC{w-guarded}) \cite{Rosati06,CaliGK13}, 
\emph{sticky}, \emph{sticky-join}, \emph{weakly-sticky-join} (\SC{wsj}) \cite{CaliGP12,GogaczM17},
\emph{shy} (\SC{shy}) \cite{LeoneMTV12} and 
\emph{weakly-recursive} (\SC{wr}) \cite{wr-2012}.
The \emph{weakly-recursive} class
is only defined for the so-called \emph{simple} TGDs, which are TGDs where the variables are only allowed
to occur once in each atom and each atom do not mention constants \cite{wr-2012}.

Another research direction
that sprangs up from those previously identified classes is the possibility of obtaining more 
expressive languages by a direct combination (i.e., union) of those classes, e.g., see
\cite{KrotzschR11,CaliGP12,g13,GottlobMP13}. A major challenge then in this 
direction is that the union of two decidable classes is not necessarily decidable \cite{BagetLMS11},
e.g., it has been shown in \cite{GottlobMP13} that the union of the classes linear and sticky is 
undecidable. 

At a model theoretic level, the results in \cite{Rosati06} and \cite{BaranyGO10} 
had respectively shown that the finite model property holds for the linear and 
guarded fragments of TGDs. It is folklore that a class of \emph{first-order} (FO) theories ${\cal C}$ 
is said to have the \emph{finite model} (FM) property if $\phi$ $\in$ ${\cal C}$ satisfiable iff
$\phi$ has a finite model.  
The recent work in \cite{GogaczM17} had further extended the 
result in \cite{Rosati06} for linear TGDs into the sticky-join TGDs. As will be revealed 
from this paper, our work further generalizes these previous results.


Despite these efforts, there are still some examples of simple 
TGDs that do not fall under the aforementioned classes. 
\begin{example}\label{motive_examp}
	Let $\tgds_1$ be a set of TGDS comprising of the following rules:	

	\vspace*{-0.50cm}

	\begin{align}
		\tgd_{11}:&\,\,\,\SF{t}(X,Y)\wedge\ra\exists Z\,\SF{t}(Y,Z)\wedge\SF{u}(Y,Z),\label{TGD_examp_rule_1}\\
		\tgd_{12}:&\,\,\,\SF{t}(X,Y)\wedge\SF{u}(Y,Z)\ra\SF{t}(Y,Z)\wedge\SF{u}(X,Y).\label{TGD_examp_rule_2}
	\end{align}

	\vspace*{-0.1cm}

\end{example}

\noindent
Then it can be checked that $\tgds_1$ does not fall into any of the classes previously mentioned above,
and neither is it \emph{glut-guarded} (\SC{g-guarded}) \cite{KrotzschR11} nor 
\emph{tame} (\SC{tame}) \cite{GottlobMP13}.
On the other hand, because none of the head atoms ``$\SF{t}(Y,Z)$" and ``$\SF{u}(X,Y)$" of $\tgd_{12}$ mentions
the two cyclically-affected body variables ``$X$" and ``$Z$" together
(which is under some pattern that we will generalize in Section 3), 
then it can be shown that for any database $D$ and query $Q$, 
it is sufficient to only consider a finite number of \emph{labeled nulls} in $\CHASE(D,\tgds_1)$
to determine if $\CHASE(D,\tgds_1)$ $\models$ $Q$.
Actually, $\tgds_1$ falls under the new class of TGDs we call \textit{triangularly-guarded}, which is a new class of 
TGDs that strictly contains several of the main syntactic classes, including $\SC{wa}$, $\SC{w-guarded}$, 
$\SC{wsj}$, $\SC{g-guarded}$, $\SC{shy}$, $\SC{tame}$ and $\SC{wr}$.

The rest of the paper is structured into three main parts as follows: Section 2 provides background notions and definitions
about databases, TGDs and the problem of (boolean) conjunctive query answering; Section 3 introduces 
the triangularly-guarded $\TG$ class of TGDs; while Section 4 looks at the main results and shows 
that $\TG$ is both decidable and strictly contains some of the main syntactic classes mentioned above, and 
also concludes the paper with some remarks.

\section{Preliminaries}\label{prelim}

\subsection{Basic notions and notations}\label{prelim_rel_TGDs}

We assume three countably infinite pairwise disjoint sets $\Gamma_{\cal V}$, $\Gamma_{\cal C}$ and $\Gamma_{\sf N}$ of 
\textit{variables}, \textit{constants} and \textit{labeled nulls}, respectively. 
We further assume that $\Gamma_{\cal V}$ is partitioned into two disjoint sets
$\Gamma_{\cal V}^\forall$ and $\Gamma_{\cal V}^\exists$ 
(i.e., $\Gamma_{\cal V}$ $=$ $\Gamma_{\cal V}^\forall$ $\cup$ $\Gamma_{\cal V}^\exists$), where
$\Gamma_{\cal V}^\forall$ and $\Gamma_{\cal V}^\exists$ denote the sets
of \emph{universally} ($\forall$) and \emph{existentially} ($\exists$) quantified variables, respectively.
We also assume that the set of labeled nulls $\Gamma_{\sf N}$ contains elements of the form
$\{{\sf n}_i$ $\mid$ $i$ $\in$ $\mathbb{N}\}$, where $\mathbb{N}$ is the set of natural numbers.
Intuitively, $\Gamma_{\sf N}$ is the set of ``fresh" Skolem terms that are disjoint from the set of
constants $\Gamma_{\cal C}$. 

A \textit{relational schema} ${\cal R}$ (or just \textit{schema}) is a set of 
\textit{relational symbols} (or \textit{predicates}), where each is associated with some
number $n$ $\geq$ $0$ called its \textit{arity}. We denote by $r/n$ as the relational 
symbol $r$ $\in$ ${\cal R}$ whose arity is $n$, and by $|r|$ as the arity of $r$, i.e.,
$|r|$ $=$ $n$. We further denote by $r[i]$ as the $i$-th \textit{argument} 
(or \textit{attribute}) of $r$ where $i$ $\in$ $\big\{0$,$\ldots$,$|r|\big\}$. 
We denote by $\ARG(r)$ as the set of arguments 
$\big\{r[i]$ $\mid$ $i$ $\in$ $\{0$,$\ldots$,$|r|\}\big\}$ of $r$. We 
extend this notion to the set of relational symbols ${\cal R}$, i.e.,
$\ARG({\cal R})$ $=$ $\bigcup_{r\in{\cal R}}\ARG(r)$.

A \textit{term} $t$ is any element from the set $\Gamma_{\cal V}$ $\cup$ $\Gamma_{\cal C}$ $\cup$ $\Gamma_{\sf N}$.
Then an \textit{atom} $\atoma$ is a construct of the form $r(t_1,\ldots,t_n)$ such that: 
(1) $r$ $\in$ ${\cal R}$; (2) $n$ $=$ $|r|$; and
(3) $t_i$ \big(for $i$ $\in$ $\{1,\ldots,n\}$\big) is a term. We denote tuples of atoms by 
$\vatoma$, e.g., $\vatoma$ $=$ $\atoma_1\ldots\atoma_l$, $\vatomb$ $=$ $\atoma_1\ldots\atoma_m$, 
$\vatomc$ $=$ $\atomc_1\ldots\atomc_n$, etc., and its length by $|\vatoma|$.

We denote by $\REL(\atoma)$, $\TERMS(\atoma)$,
$\VAR(\atoma)$, $\CONST(\atoma)$ and $\NULLS(\atoma)$ as the relational symbol, the set of terms, variables, constants
and labeled nulls mentioned in atom $\atoma$, respectively. We extend these notions to a set or tuples of atoms
$S$ such that  $\TERMS(\atoma)$, $\VAR(S)$, $\CONST(S)$ and $\NULLS(S)$ denotes the sets 
$\bigcup_{\atoma\in S}\TERMS(\atoma)$, $\bigcup_{\atoma\in S}\VAR(\atoma)$, $\bigcup_{\atoma\in S}\CONST(\atoma)$ 
and  $\bigcup_{\atoma\in S}\NULLS(\atoma)$, respectively. 
We say that a tuple of atoms
$\vatoma$ $=$ $\atoma_1\ldots\atoma_l$ is \emph{connected} if either: (1) $\vatoma$ is an atom, or 
(2) $\TERMS(\atoma_i)$ $\cap$ $\TERMS(\atoma_{i+1})$ $\neq$ $\emptyset$ holds, for each $i$ $\in$ $\{1,\ldots,l-1\}$.

An \emph{instance} $I$ is 
any set (can be infinite) of atoms such that $\VAR(I)$ $=$ $\emptyset$, i.e., contains no variables. 
A \emph{database} $D$ is a finite set of ground atoms 
$\VAR(D)$ $=$ $\emptyset$ and $\NULLS(D)$ $=$ $\emptyset$.

Given an atom $\atoma$ $=$ $r(t_1,\ldots,t_n)$, we denote by $\ARG(\atoma)$ as the set of arguments
$\big\{r[i]$ $\mid$ $i\in\{1,\ldots,n\}\big\}$.
For a tuple of variables $\vect{X}$ and atom $\atoma$ $=$ $r(t_1,\ldots,t_n)$,
we denote by $\ARG(\atoma)\REST_{\vect{X}}$ as the set of arguments 
$\big\{r[i]$ $\mid$ $i$ $\in$ $\{1,\ldots,n\}$ and $t_i$ $=$ $X\big\}$,
i.e., the set of arguments
in $\ARG(\atoma)$ but \emph{restricted} to those mentioned variables from $\vect{X}$.
Symmetrically, using similar notions to the ``$\ARG$" concept just previously mentioned above, 
for a given atom $\atoma$ $=$ $r(t_1,\ldots,t_n)$ and set of arguments $A$ $\subseteq$ $\ARG(\atoma)$,
we denote by $\VAR(\atoma)\REST_A$ as the set of variables 
$\big\{X$ $\mid$ $t_i$ $=$ $X$ and $r[i]$ $\in$ $A\big\}$,
i.e., the set of all the variables mentioned in $\atoma$ but restricted to
those appearing in argument positions from $A$.

Given two sets of terms $T_1$ and $T_2$, an assignment $\theta:$ $T_1$ $\longrightarrow$ $T_2$
is a \emph{function} from $T_1$ onto $T_2$ such that 
$t$ $\in$ $(T_1\cap\Gamma_{\cal C})$ implies $\theta(t)$ $=$ $t$, 
i.e., identity for the constants $\Gamma_{\cal C}$. 
Then for a given atom $\atoma$ $=$ $r(t_1,\ldots,t_n)$,
a set of terms $T$ and an assignment $\theta:$ $\VAR(\atoma)$ $\longrightarrow$ $T$,
a \textit{substitution of $\atoma$ under $\theta$}  \big(or just \textit{substitution} for convenience\big), 
denoted $\atoma\theta$ \big(or sometimes $\theta(\atoma)$\big), 
is the atom such that $\atoma\theta$ $=$ $r\big(\theta(t_1),\ldots,\theta(t_n)\big)$. 
We naturally extend
to conjunctions of atoms $\atoma_1$ $\wedge$ $\ldots$ $\wedge$ $\atoma_n$ so that
$\theta(\atoma_1$ $\wedge$ $\ldots$ $\wedge$ $\atoma_n)$ $=$ 
$\atoma_1\theta$ $\wedge$ $\ldots$ $\wedge$ $\atoma_n\theta$.
Given two assignments $\theta_1:$ $T_1$ $\longrightarrow$ $T_2$ and $\theta_2:$ $T_2$ $\longrightarrow$ $T_3$,
we denote by $\theta_2\circ\theta_1$ as the \textit{composition} of $\theta_1$ with $\theta_2$ such that  
$\theta_2\circ\theta_1:$ $T_1$ $\longrightarrow$ $T_3$ and 
$(\theta_2\circ\theta_1)(t)$ $=$ $\theta_2\big(\theta_1(t)\big)$, for all $t$ $\in$ $T_1$.
Then lastly, given again an assignment $\theta:$ $T_1$ $\longrightarrow$ $T_2$ and some
set of terms $T'$ $\subseteq$ $T_1$, we denote by $\theta\REST_{T'}$ as the \textit{restriction}
of the assignment $\theta$ to the domain $T'$ $\subseteq$ $T_1$ such that 
$\theta\REST_{T'}:$ $T'$ $\longrightarrow$ $T_2$, and where $\theta\REST_{T'}(t)$ $=$ $\theta(t)$
$\in$ $T_2$, for each $t$ $\in$ $T'$ $\subseteq$ $T_1$.

\subsection{TGDs, BCQ-Ans and Chase}

A \textit{tuple generating dependency} (TGD) rule $\tgd$ of schema ${\cal R}$
is a \textit{first-order} (FO) formula of the form \cite{BeeriV81}:
\vspace*{-0.4cm}
\begin{align}
	\forall\vect{X}\vect{Y}\big(\Phi(\vect{X},\vect{Y})\,\ra\,\exists\vect{Z} \Psi(\vect{Y},\vect{Z})\big),\label{tgd_rule}
\end{align}
\vspace*{-0.6cm}

\noindent 
where:
\begin{itemize}
	\item $\vect{X}$ $=$ $ X_1\ldots X_k$, $\vect{Y}$ $=$ $ Y_1\ldots Y_l$ and 
	      $\vect{Z}$ $=$ $ Z_1\ldots Z_m$ are pairwise disjoint tuple of variables, and where they are called
	      the \textit{local}, \textit{shared} and \textit{exitential} variables, respectively, and thus, we assume that
	      $\vect{X}\vect{Y}$ $\subseteq$ $\Gamma_{\cal V}^\forall$ and 
	      $\vect{Z}$ $\subseteq$ $\Gamma_{\cal V}^\exists$;
    \item $\Phi(\vect{X},\vect{Y})$ $=$ $b_1(\vect{V}_{1})\wedge\ldots\wedge b_n(\vect{V}_{n})$ is a conjunction
          of atoms such that $\vect{V}_i$ $\subseteq$ $\vect{X}\vect{Y}$ and $b_i$ $\in$ ${\cal R}$, 
          for $i$ $\in$ $\{1$,$\ldots$,$n\}$;
    \item $\Psi(\vect{Y},\vect{Z})$ $=$ $r_1(\vect{W}_1)\wedge\ldots\wedge r_m(\vect{W}_m)$ is a conjunction 
          of atoms where $\vect{W}_i$ $\subseteq$ $\vect{Y}\vect{Z}$ and
          $r_i$ $\in$ ${\cal R}$, for $i$ $\in$ $\{1$,$\ldots$,$m\}$.
\end{itemize}
For a given TGD $\tgd$ of the form (\ref{tgd_rule}), we denote by 
$\BD(\tgd)$ as the set  of atoms $\big\{b_1(\vect{V}_{1})$,$\ldots$,$b_n(\vect{V}_{n})\big\}$, which
we also refer to as the \textit{body} of $\tgd$. Similarly, by $\HD(\tgd)$ we denote the
set of atoms $\big\{r_1(\vect{W}_1),\ldots,r_m(\vect{W}_m)\big\}$, 
which we also refer to as the \textit{head}
of $\tgd$. For convenience, when it is clear from the context, we simply drop the quantifiers in
(\ref{tgd_rule}) such that a TGD rule $\tgd$ of the form (\ref{tgd_rule}) can simply be referred to as: 
$\Phi(\vect{X},\vect{Y})$ $\ra$ $\Psi(\vect{Y},\vect{Z})$. Then, for a given set of 
TGDs $\tgds$, we denote by $\ATOMS(\tgds)$ as the set of all atoms occurring in $\tgds$
such that $\ATOMS(\tgds)$ $=$ $\bigcup_{\tgd\in\tgds}\big(\BD(\tgd)\cup \HD(\tgd)\big)$, and by
$\REL(\tgds)$ as the set of all relational symbols mentioned in $\tgds$. Then lastly,
for convenience later on, for a given rule $\tgd$ of the form (\ref{tgd_rule}) and
atom $\atoma$ $\in$ $\HD(\tgd)$, we denote by $\UVAR(\atoma)$ 
and $\EVAR(\atoma)$ as the set of variables $\VAR(\atoma)\cap\vect{Y}$ and 
$\VAR(\atoma)\cap\vect{Z}$, respectively, i.e., the set of all the universally ($\forall$) 
and existentially ($\exists$) quantified variables of $\atoma$, respectively. 
We extend this notion to the TGD rule $\tgd$ of the form (\ref{tgd_rule}) 
so that we set $\VAR(\tgd)$ $=$ $\vect{XYZ}$, $\UVAR(\tgd)$ $=$ $\vect{XY}$ and 
$\EVAR(\tgd)$ $=$ $\vect{Z}$.

A \emph{boolean conjunctive query} (BCQ) $Q$ is a FO formula $\exists\vect{X}\varphi(\vect{X})$
$\ra$ $q$ such that $\varphi(\vect{X})$ $=$ $r_1(\vect{Y}_1)\wedge\ldots\wedge r_l(\vect{Y}_l)$,
where $r_i$ $\in$ ${\cal R}$ and $\vect{Y}_i$ $\subseteq$ $\vect{X}$, 
for each $i$ $\in$ $\{1$, $\ldots$, $l\}$, and where we set 
$\BD(Q)$ $=$ $\{r_1(\vect{Y}_1),\ldots,r_l(\vect{Y}_l)\}$. 
Given a database $D$ and a set of TGDs $\tgds$,
we say that $D$ $\cup$ $\tgds$ \emph{entails} $Q$, denoted $D$ $\cup$ $\tgds$ $\models$ $Q$,
iff $D$ $\cup$ $\tgds$ $\models$ $\exists\vect{X}\varphi(\vect{X})$. The central problem tackled 
in this work is the \emph{boolean conjunctive query answering} (BCQ-Ans): 
\emph{given a database $D$, a set of TGDs $\tgds$ and BCQ $Q$, does $D$ $\cup$ $\tgds$ $\models$ $Q$?}
It is well know that BCQ-Ans is \emph{undecidable} in general \cite{BeeriV81}.

The \emph{chase procedure} (or just \emph{chase}) 
\cite{MaierMS79,JohnsonK84,AbiteboulHV95,FaginKMP05,DeutschNR08,ZhangZY15} 
is a main algorithmic tool proposed for checking 
implication dependencies \cite{MaierMS79}. For an instance $I$, assignment $\eta$ and TGD 
$\tgd$ $=$ $\Phi(\vect{X},\vect{Y})$ $\ra$ $\Psi(\vect{Y},\vect{Z})$, 
we have that $I\xrightarrow{\sigma,\,\eta} I'$ defines a single \emph{chase step} as follows:
$I'$ $=$ $I$ $\cup$ $\big\{\eta'\big(\Psi(\vect{Y},\vect{Z})\big)\big\}$ such that: (1)
$\eta:$ $\vect{X}\vect{Y}$ $\longrightarrow$ $\Gamma_{\cal C}$ $\cup$ $\Gamma_{\SF{N}}$ and
$\eta\big(\Phi(\vect{X},\vect{Y})\big)$  $\subseteq$ $I$; and 
(2) $\eta':$ $\vect{X}\vect{Y}\vect{Z}$ $\longrightarrow$ $\Gamma_{\cal C}$ $\cup$ $\Gamma_{\SF{N}}$ 
and $\eta'\REST_{\vect{XY}}$ $=$ $\eta$. As in the literatures, we further assume here that
each labeled nulls used to eliminate the $\exists$-quantified variables 
in $\vect{Z}$ follows lexicographically all the previous ones, i.e.,
follows the order $\SF{n}_i$, $\SF{n}_{i+1}$, $\SF{n}_{i+2}$, $\ldots$, etc. 
A \emph{chase sequence} of a database $D$ w.r.t. to a set of TGDs $\tgds$ is a sequence of chase
steps $I_i\xrightarrow{\sigma_i,\,\eta_i} I_{i+1}$, where $i$ $\geq$ $0$, $I_0$ $=$ $D$ and
$\sigma_i$ $\in$ $\tgds$. 
An infinite chase sequence $I_i\xrightarrow{\sigma_i,\,\eta_i} I_{i+1}$
is \emph{fair} if $\eta\big(\Phi(\vect{X},\vect{Y})\big)$ $\subseteq$ $I_i$,
for some $\eta:$ $\vect{X}\vect{Y}$ $\longrightarrow$ $\Gamma_{\cal C}$ $\cup$ $\Gamma_{\SF{N}}$
and $\tgd$ $=$ $\Phi(\vect{X},\vect{Y})$ $\ra$ $\Psi(\vect{Y},\vect{Z})$ $\in$ $\tgds$,
implies
$\exists\eta':$ $\vect{X}\vect{Y}\vect{Z}$ $\longrightarrow$ $\Gamma_{\cal C}$ $\cup$ $\Gamma_{\SF{N}}$, 
where $\eta'\REST_{\vect{X}\vect{Y}}$ $=$ $\eta$, such that $\eta'\big(\Psi(\vect{Y},\vect{Z})\big)$ 
$\subseteq$ $I_k$ and $k$ $>$ $i$.  
Then finally, we let 
$\CHASE(D,\tgds)$ $=$ $\bigcup_{i=0}^{\infty}I_i$. 
\begin{theorem}\cite{DeutschNR08,CaliGP12}
	Given a database $D$, TGDs $\tgds$ and BCQ $Q$, $D$ $\cup$ $\tgds$ $\models$ $Q$ iff
	$\CHASE(D,\tgds)$ $\models$ $Q$.	
\end{theorem}

\subsection{Cyclically-affected arguments}\label{prelim_AG}

As observed in \cite{LeoneMTV12}, the notion of \emph{affected arguments} in \cite{CaliGK13} can 
sometimes consider arguments that may not actually admit a ``firing" mapping $\forall$-variables into nulls.
For this reason, it was introduced in \cite{LeoneMTV12} the notion of a ``null-set." 
Given a set of TGDs $\tgds$,
let $\atoma$ $\in$ $\ATOMS(\tgds)$, $a$ $\in$ $\ARG(\atoma)$ and $X$ $=$ $\VAR(\atoma)\REST_{a}$. 
Then the \emph{null-set} of $a$ in $\atoma$ under $\tgds$, denoted as
$\NULLSET(a,\atoma,\tgds)$ \big(or just $\NULLSET(a,\atoma)$ if clear from the context\big), 
is defined inductively as follows: If $\atoma$ $\in$ $\HD(\tgd)$,
for some $\tgd$ $\in$ $\tgds$, then: 
(1) $\NULLSET(a,\atoma)$ $=$ $\{\SF{n}_X^\tgd\}$,
\footnote{We assume that $(\tgd,X)$ $\neq$ $(\tgd',X')$ implies 
          $\SF{n}_X^\tgd$ $\neq$ $\SF{n}_{X'}^{\tgd'}$,
          for each pair of elements $(\SF{n}_X^\tgd$, $\SF{n}_{X'}^{\tgd'})$ of all null-sets
          \cite{LeoneMTV12}.}
if $\VAR(\atoma)\REST_{a}$ $=$ $X$ $\in$ $\EVAR(\tgd)$, or 
(2) $\NULLSET(a,\atoma)$ is the intersection of all null-sets $\NULLSET(b,\atomb)$ such that
$\atomb$ $\in$ $\BD(\tgd)$, $b$ $\in$ $\ARG(\atomb)$ and $\VAR(\atomb)\REST_{b}$ $=$ $X$.
Otherwise, if $\atoma$ $\in$ $\BD(\tgd)$, for some $\tgd$ $\in$ $\tgds$, then
$\NULLSET(a,\atoma)$ is the union of all $\NULLSET(a,\vect{a}')$ such that
$\REL(\vect{a}')$ $=$ $\REL(\vect{a})$ and $\vect{a'}$ $\in$ $\HD(\tgd')$, where
$\tgd'$ $\in$ $\tgds$.

Borrowing similar notions from \cite{KrotzschR11} used in the identification of the so-called 
\emph{glut variables}, the \textit{existential dependency graph} ${\cal G}_{\exists}(\tgds)$ 
is a graph $(N,E)$, whose nodes $N$ is the union of all $\NULLSET(a,\atoma)$, where 
$\atoma\in\ATOMS(\tgds)$ and $a\in\ARG(\atoma)$, and edges:

\vspace*{-0.3cm}		

\begin{align}	
	&\hspace*{-1.3cm}E=\big\{\,\big(\SF{n}_{Z'}^{\tgd'},\SF{n}_{Z}^{\tgd}\big)\,\mid\,
	         \exists\tgd\in\tgds\mbox{ of form (\ref{tgd_rule})},\,\exists Y\in\vect{Y},\nonumber
\end{align}

\vspace*{-0.4cm}

\begin{align}	         
	      &\SF{n}_{Z'}^{\tgd'}\in\,\bigcap\NULLSET(Y,\tgd,\tgds)
	      \mbox{ and }\NULLSET(a,\atoma)=\{\SF{n}_{Z}^{\tgd}\},\nonumber
\end{align}

\vspace*{-0.4cm}

\begin{align}	      
	      &\hspace*{-3.5cm}\mbox{for some }\atoma\in\HD(\tgd)\mbox{ and }Z\in\vect{Z}\,\big\},\nonumber	
\end{align}


%
\noindent
where $\bigcap\NULLSET(Y,\tgd,\tgds)$ denotes the intersection of all $\NULLSET(b,\atomb,\tgds)$ 
such that $\atomb\in\BD(\tgd)$, $b\in\ARG(\atomb)$ and $Y=\VAR(\atomb)\,\REST_b$.
We note that our definition of a dependency graph here generalizes the \emph{existential dependency graph}
in \cite{KrotzschR11} by combining the notion of null-sets in \cite{LeoneMTV12}. 
Then with the graph ${\cal G}_{\exists}(\tgds)$ $=$ $(N,E)$ as defined above, 
we denote by $\CYCNULL(\tgds)$ as the smallest subset of $N$ such that 
$\SF{n}^{\tgd}_{Z}$ $\in$ $\CYCNULL(\tgds)$ iff either: (1) $\SF{n}^{\tgd}_{Z}$ is in a cycle in ${\cal G}_{\exists}(\tgds)$,
or (2) $\SF{n}^{\tgd}_{Z}$ is reachable from some other node $\SF{n}^{\tgd'}_{Z'}$ $\in$ $\CYCNULL(\tgds)$,
where $\SF{n}^{\tgd'}_{Z'}$ is in a cycle in ${\cal G}_{\exists}(\tgds)$.

\section{Triangularly-Guarded (TG) TGDs}

This section now introduces the triangularly-guarded class of TGDs, which is the focus of 
this paper. We begin with an instance of a BCQ-Ans problem that corresponds to 
a need of an infinite number of labeled nulls in the underlying chase derivation.

\begin{example}[\textbf{\textit{Unbounded nulls}}]\label{cyc_aag_examp}
	Let $\tgds_2$ $=$ $\{\tgd_{11},\tgd^*_{12},\}$ be the set of TGDs obtained from 
	$\tgds_1$ $=$ $\{\tgd_{11},\tgd_{12}\}$ of
	Example \ref{motive_examp} by just changing the rule $\tgd_{12}$ into the rule $\tgd^*_{12}$ such that:
	
	\vspace*{-0.4cm}	

	\begin{align}
		&\tgd^*_{12}:\,\,\SF{t}(X,Y)\wedge\SF{u}(Y,Z)\ra\SF{t}(X,Z)\wedge\SF{u}(X,Y).
		\label{cyc_aag_examp_form_1}
	\end{align}

	\vspace*{-0.1cm}
	
	\noindent
	Then we have that $\tgd^*_{12}$ of $\tgds_2$ above is obtained from $\tgd_{12}$ of $\tgds_1$ by
	changing the variable ``$Y$" in the head atom ``$\SF{t}(Y,Z)$" of $\tgd_{12}$ into ``$X$", 
	i.e., to  obtain ``$\SF{t}(X,Z)$". Intuitively,
	this allows the two variables ``$X$" and ``$Y$" to act as place holders that combines labeled nulls together
	in the head atom ``$\SF{t}(X,Z)$" of $\tgds^*_{12}$.
	Now let $D_2$ $=$ $\big\{\SF{t}(c_1,c_2),\SF{u}(c_1,c_2)\big\}$ be a database, 
	where $c_1$, $c_2$ $\in$ $\Gamma_{\cal C}$ and $c_1$ $\neq$ $c_2$, and $Q_2$ the BCQ 
	$\exists X\SF{t}(X,X)\ra q$. Then 
	we have that $D_2$ $\cup$ $\tgds_2$ $\cup$ $\{\neg\exists X\SF{t}(X,X)\}$
	$\equiv$
	$D_2$ $\cup$ $\tgds_2$ $\cup$ $\{\forall X\neg\SF{t}(X,X)\}$ can only be satisfied 
	by the infinite model $M$ of the form $M$ $=$
	$D_2$ $\cup$ $\bigcup_{1\,\leq\,i\,<\,j}\big\{\SF{t}(c_i,c_j)\big\}$, where 
	we assume $i$ $\neq$ $j$ implies $c_i$ $\neq$ $c_j$. 
	Therefore, since $D_2$ $\cup$ $\tgds_2$ $\cup$ $\{\forall X\neg\SF{t}(X,X)\}$ is satisfiable,
	(albeit infinitely), it follows that $D_2$ $\cup$ $\tgds_2$ $\not\models$ $Q_2$.
\end{example}

In database $D_2$ and set TGDs $\tgds_2$ $=$ $\{\tgd_{11}$, $\tgd^*_{12}\}$ of Example \ref{cyc_aag_examp}, 
we get from $\tgd_{11}$ the sequence of atoms 
$\SF{t}(c_2,\SF{n}_1)$, $\SF{t}(\SF{n}_1,\SF{n}_2)$, $\SF{t}(\SF{n}_2,\SF{n}_3)$, 
$\ldots$ $\SF{t}(\SF{n}_{k-1},\SF{n}_k)$ $\in$ $\CHASE(D_2,\tgds_2)$, where $\SF{n}_i$ $\in$ $\Gamma_{\SF{N}}$,
for each $i$ $\in$ $\{1,\ldots,k\}$. Moreover, by the repeated applications of
$\tgd^*_{12}$, we further get that $\SF{t}(\SF{n}_i,\SF{n}_k)$ $\in$ $\CHASE(D_2,\tgds_2)$,
for each $i$ $\in$ $\{1,\ldots,k-1\}$, i.e., $\SF{n}_i$ and $\SF{n}_k$, for each $i$ $\in$ $\{1,\ldots,k-1\}$, 
will be ``pulled" together in some relation of $\SF{t}$ in $\CHASE(D_2,\tgds_2)$.
As such, for the given BCQ $Q_2$ $=$ $\exists X\SF{t}(X,X)\ra q$ also from Example \ref{cyc_aag_examp},
since $D_2\cup\tgds_2$ $\models$ $Q_2$ iff $D_2\cup\tgds_2\cup\{\forall X\neg\SF{t}(X,X)\}$
is not satisfiable, then the fact that we have to satisfy the literal ``$\neg\SF{t}(X,X)$" for all ``$X$",
and because $\SF{t}(\SF{n}_i,\SF{n}_k)$ $\in$ $\CHASE(D_2,\tgds_2)$, for $i$ $\in$ $\{1,\ldots,k-1\}$,  
implies that each of the $\SF{n}_i$ must be of different values from $\SF{n}_k$, and thus 
cannot be represented by a finite number of distinct labeled nulls.

\subsection{Triangular-components of TGD extensions}

In contrast to $\tgds_2$ from Example \ref{cyc_aag_examp},
what we aim to achieve now is to identify syntactic conditions on 
TGDs so that such a ``distinguishable relation" of labeled nulls is limited in the chase derivation. 
As a consequence,
we end up with some nulls that need not be distinguishable from another, and as such, we can actually
{\em re-use} these nulls without introducing new ones in the chase derivation. This leads BCQ-Ans to be decidable.

\begin{definition}[\textbf{\textit{TGD extension}}]\label{tgd_extension}
	Given a set of TGDs $\tgds$, we denote by $\tgds^+$ as the \emph{extension} of $\tgds$, 
	and is inductively defined as follows:
	
	\vspace*{-0.4cm}

	\begin{align}
		&\hspace*{-3.4cm}\tgds^0=\big\{\,\big\LA\BD(\tgd),\HD(\tgd)\big\RA\,\mid\,
		\tgd\in\tgds\,\big\};\label{tgds_ext_def_base}
	\end{align}

	\vspace*{-0.6cm}	

	\begin{align}
		&\hspace*{-3.6cm}\tgds^{i+1}=\,\tgds^i\,\cup\hspace*{3cm}\label{tgds_ext_def_ind_1}
    \end{align}	

	\vspace*{-0.6cm}			
		
	\begin{align}		
		\big\{\,&\big\LA B_1\eta_1\cup B^*,H_2(\eta_2\circ\theta)\big\RA\mbox{ \large $\mid$ }\LA B_1,H_1\RA\in\tgds^i,\,
		       \LA B_2,H_2\RA\in\tgds^i\nonumber\\
	    &\mbox{and where }
		\eta_1:\VAR(B_1\cup H_1)\longrightarrow\TERMS(B_1\cup H_1)\mbox{ and}
		\nonumber\\		
		&\eta_2:\VAR(B_2\theta\cup H_2\theta)\longrightarrow\TERMS(B_1\eta_1\cup B_2\theta\cup H_2\theta)
		\nonumber\\
		&\mbox{ such that:}\nonumber\\
		&\mbox{(1) }\theta\mbox{ is a renaming (bijective) substitution such that }\nonumber\\
		&\hspace*{0.5cm}\VAR(B_1\cup H_1)\cap\VAR(B_2\theta\cup H_2\theta)=\emptyset;\nonumber\\	
		&\mbox{(2) }\exists H'_1\subseteq H_1\mbox{ and }\exists B'_2\subseteq B_2\theta
		                 \mbox{ such that }H'_1\eta_1=B'_2\eta_2\nonumber\\
		&\hspace*{0.5cm}\mbox{corresponds to the MGU of }H'_1\mbox{ and }B'_2,\mbox{ and}\nonumber\\			
		&\hspace*{0.5cm}\eta_1(X)=X,\mbox{ for each }X\in\VAR(B_1)\setminus\VAR(H'_1);\nonumber\\		
		&\mbox{(3) }B^*= B_2(\eta_2\circ\theta)\setminus B'_2\eta_2\,\big\}.
		\hspace*{-0.5cm}\label{tgds_ext_def_ind_2}
 \end{align}

	\vspace*{-0.2cm}

\end{definition}
\noindent
Then we set $\tgds^+$ $=$ $\tgds^\infty$ as the fixpoint of $\tgds^i$. We note that even though
$\tgds^+$ can be infinite in general, it follows from Theorem \ref{membership_complexity} that it
is enough to consider a finite number of iterations $\tgds^i$ to determine ``recursive triangular-components"
(as will be defined exactly in Definition \ref{triangular_comp_def}).

\comment
{
To keep the computation of $\tgds^+$ finite, we further assume that for each
$\LA B_1,H_1\RA$, $\LA B_2,H_2\RA$ $\in$ $\tgds^+$, $\LA B_1,H_1\RA$ ``homomorphic" to $\LA B_2,H_2\RA$
implies that $\LA B_1,H_1\RA$ $=$ $\LA B_2,H_2\RA$. Here, we say that  
$\LA B_1,H_1\RA$ \emph{homomorphic} $\LA B_2,H_2\RA$ $\in$ $\tgds^+$ if there exists a 
function $\theta:$ $\VAR(B_1\cup H_1)$ $\longrightarrow$ $\VAR(B_2\cup H_2)$ 
such that: 
(1) $\LA B_1\theta,H_1\theta\RA$ $=$ $\LA B_2,H_2\RA$; and 
(2) $r(t_1,\ldots,t_k)$ $\in$ $B_1$ \big(resp. $r(t_1,\ldots,t_k)$ $\in$ $H_1$\big) iff
$r(\theta(t_1),\ldots,\theta(t_k))$ $\in$ $B_2$ \big(resp. $r(\theta(t_1),\ldots,\theta(t_k))$ $\in$ $H_2$\big).
}

The TGD extension $\tgds^+$ of $\tgds$ contains as elements pairs of sets of atoms of the form 
``$\LA B,H\RA$". Loosely speaking, the set $B$ represents the body of some TGD in $\tgds$ while
$H$ as the head atoms that can be linked (transitive) through the repeated applications of the steps in
(\ref{tgds_ext_def_ind_1})-(\ref{tgds_ext_def_ind_2}) 
(which is done until a fixpoint is reached).
The base case $\tgds^0$ in (\ref{tgds_ext_def_base}) first considers the 
pairs $\LA B,H\RA$, where $B$ $=$ $\BD(\tgd)$ and $H$ $=$ $\HD(\tgd)$, for each $\tgd$ $\in$ $\tgds$.
Inductively, assuming we have already computed $\tgds^{i}$, we have that $\tgds^{i+1}$ is obtained by
adding the previous step $\tgds^{i}$ as well as adding the set 
as defined through (\ref{tgds_ext_def_ind_1})-(\ref{tgds_ext_def_ind_2}).

More specifically, using similar ideas to the \emph{TGD expansion} in \cite{CaliGP12} that was used in 
identifying the \emph{sticky-join} class of TGDs and \emph{tame reachability} in \cite{GottlobMP13}
used for the \emph{tame} class, 
the set (\ref{tgds_ext_def_ind_1})-(\ref{tgds_ext_def_ind_2}) considers the other head types
that can be (transitively) reached from some originating TGD. Indeed, as described in (\ref{tgds_ext_def_ind_1})-(\ref{tgds_ext_def_ind_2}), for $\LA B_1,H_1\RA$ $\in$ $\tgds^i$ and 
$\LA B_2,H_2\RA$ $\in$ $\tgds^{i}$, 
we add the pair $\LA B_1\eta_1\cup B^*,H_2(\eta_2\circ\theta)\RA$ into 
$\tgds^{i+1}$. Intuitively, with the assignment ``$\eta_2\circ\theta$" as described in (\ref{tgds_ext_def_ind_1})-(\ref{tgds_ext_def_ind_2}), the aformentioned pair $\LA B_1\eta_1\cup B^*,H_2(\eta_2\circ\theta)\RA$
encodes the possibility that ``$H_2(\eta_2\circ\theta)$" can be derived transitively from bodies $B_1\eta_1$ and 
$B^*$ $=$ $B_2(\eta_2\circ\theta)\setminus B'_2\eta_2$.
We note that the renaming function ``$\theta$" is only used for pair $\LA B_2,H_2\RA$ 
in (\ref{tgds_ext_def_ind_1})-(\ref{tgds_ext_def_ind_2}) (and no renaming for pair $\LA B_1,H_1\RA$)
so that we can track some of the originating variables from $B_1$ all the way through the head
``$H_2(\eta_2\circ\theta)$", and which can be retained through iterative applications of the criterion given
in (\ref{tgds_ext_def_ind_1})-(\ref{tgds_ext_def_ind_2}). Importantly, we note that the connection
between $B_1$ and $H_2$ is inferred with $H'_1\eta_1$ $=$ $B'_2\eta_2$ 
(where $H'_1$ $\subseteq$ $H_1$ and $B'_2$ $\subseteq$ $B_2\theta$) corresponding to the 
\emph{most general unifier} (MGU) of $H'_1$ and $B'_2$ (please see Condition (2) of set (\ref{tgds_ext_def_ind_1})-(\ref{tgds_ext_def_ind_2})).

\begin{example}\label{tgd_extension_examp}
	Let $\tgds_3$ be the following set of TGD rules:

	\vspace*{-0.5cm}		    

	\begin{align}
		\tgd_{31}:&\,\,\,\SF{t}(X,Y)\ra\exists Z\,\SF{t}(Y,Z),\nonumber\\
		\tgd_{32}:&\,\,\,\SF{t}(X,Y)\ra\,\SF{s}(X)\wedge\SF{s}(Y),\nonumber\\				
		\tgd_{33}:&\,\,\,\SF{t}(X_1,V)\wedge\SF{s}(V)\wedge\SF{t}(W,Z_1)\ra\SF{u}(X_1,V,W,Z_1),\nonumber\\
		\tgd_{34}:&\,\,\,\SF{u}(X_2,Y,Y,Z_2)\ra\SF{v}(X_2,Z_2),\nonumber\\
		\tgd_{35}:&\,\,\,\SF{v}(X_3,Z_3)\ra\SF{t}(X_3,Z_3).\nonumber							
	\end{align}

	\vspace*{-0.1cm}	
	
	\noindent
	Then from the rules $\tgd_{33}$, $\tgd_{34}$ and $\tgd_{35}$, we get the three pairs
	$p_{01}$ $=$ $\LA\{\SF{t}(X_1,V)$,$\SF{s}(V)$,$\SF{t}(W,Z_1)\}$,$\{\SF{u}(X_1,V,W,Z_1)\}\RA$,
	$p_{02}$ $=$ $\LA\{\SF{u}(X_2,Y,Y,Z_2)\}$,$\{\SF{v}(X_2,Z_2)\}\RA$ and 
	$p_{03}$ $=$ $\LA\{\SF{v}(X_3,Z_3)\}$, $\{\SF{t}(X_3,Z_3)\}\RA$		
	in $\tgds^0_3$, respectively. Then through the unification of head atom
	``$\SF{u}(X_1,V,W,Z_1)$" of rule $\tgd_{33}$ with the body atom ``$\SF{u}(X_2,Y,Y,Z_2)$" of $\tgd_{34}$, 
	we get the pair $p_{11}$ $=$ $\LA\{\SF{t}(X_1,V)$, $\SF{s}(V)$,
	$\SF{t}(V,Z_1)\}$, $\{\SF{v}(X_1,Z_1)\}\RA$ $\in$ $\tgds^1_3$.
	Then finally, through the unification of the head atom ``$\SF{v}(X_2,Z_2)$" of $\tgd_{34}$ and the
	body atom ``$\SF{v}(X_3,Z_3)$" of $\tgd_{35}$, we further get the pair $p_{21}$ $=$ 
	$\LA\{\SF{t}(X_1,V)$, $\SF{s}(V)$, $\SF{t}(V,Z_1)\}$, $\{\SF{t}(X_1,Z_1)\}\RA$ $\in$ $\tgds^2_3$.
\end{example}
As will be seen in Definition \ref{triangular_comp_def}, the last pair $p_{21}$ in Example 
\ref{tgd_extension_examp} corresponds to what we will call a ``recursive triangular-component"
that will be defined precisely in Definition \ref{triangular_comp_def}.

\subsection{Triangularly-guarded TGDs}\label{triangularly-guarded_TGDs}

In this section, we now introduce the key notion of triangularly-guarded TGDs,
which are the \emph{triangular-components}. 
It will first be necessary to introduce the following notions of 
cyclically-affected only and link variables of body atoms, as well as
variable markups that borrows some concepts from \cite{CaliGP12}.

We first introduce the notion of \emph{cyclically-affected} 
only variables in the body (i.e., set $B$) of some pair $\LA B,H\RA$ $\in$ $\tgds^+$ where
$\tgds$ is a set of TGDs.
So towards this purpose, for a given pair $\LA B,H\RA$ $\in$ $\tgds^+$, we define
$\widehat{\VAR}(\tgds,B)$ (i.e., ``$\widehat{\VAR}$" is read \emph{var-hat})
as the set of variables: 
$\big\{X$ $\mid$ $X$ $\in$ $\VAR(B)$
and $\bigcap\NULLSET(X,\tgd,\tgds)[B]$ $\cap$ $\CYCNULL(\tgds)$ $\neq$ $\emptyset\big\}$,
where $\bigcap\NULLSET(X,\tgd,\tgds)[B]$ denotes the intersection of the unions 
$\bigcup_{b\in\ARG(\atomb'),\,\atomb'\in\HD(\tgd'),\,\tgd'\in\tgds}\NULLSET(b,\atomb',\tgds)$, 
for each pair $(b,\atomb)$ such that $b$ $\in$ $\ARG(\atomb)\REST_X$ and $\atomb$ $\in$ $B$.
For convenience and when clear from the context, we simply refer to $\widehat{\VAR}(\tgds,B)$
as $\widehat{\VAR}(B)$.
Intuitively, variables in $\widehat{\VAR}(B)$ are placement for which an infinite 
number of labeled nulls can possibly be propagated in the set of atoms $B$ with respect to $\tgds$.
Loosely speaking, if $B$ $=$ $\BD(\tgd)$, for some set of TGDs $\tgds$, then $\widehat{\VAR}(B)$
contains the \emph{glut-variables} in \cite{KrotzschR11} that also fails the 
\emph{shyness} property \cite{LeoneMTV12} (please see Section \ref{prelim_AG} of this paper).

Next, we introduce the link variables
between ``body atoms''. Given two atoms $\atomb_1$, $\atomb_2$ $\in$ $B$, 
for some $\LA B,H\RA$ $\in$ $\tgds^+$ where $\tgds$ is a set of TGDs, 
we set $\LINK(\tgds^+,B,\atomb_1,\atomb_2)$
\big(or just $\LINK(B,\atomb_1,\atomb_2)$ when clear from the context\big) as the set of variables
in the intersections 
$\big(\VAR(\atomb_1)$ $\cap$ $\VAR(\atomb_2)\big)$ $\cap$ $\widehat{\VAR}(\tgds,B)$.
Intuitively, $\LINK(B,\atomb_1,\atomb_2)$ denotes the cyclically-affected only variables
of $B$ that can actually ``join" (link) two common nulls between the body atoms $\atomb_1$ and $\atomb_2$ 
that can be obtained through some firing substitution.

Lastly, we now introduce the notion of variable markup.
Let $\atoma$, $\atomc$ and $\atoma'$ be three atoms such that $\REL(\atoma)$ $=$ $\REL(\atoma')$. 	 
Then similarly to \cite{CaliGP12}, we define the ``markup procedure" as follows. 
For the base case, we let $\atoma^0$ (resp. $\atomc^0$) denote the atom obtained from 
$\atoma$ (resp. $\atomc$) by marking each variable 
$X$ $\in$ $\VAR(\atoma)$ (resp. $X$ $\in$ $\VAR(\atomc)$) such that
$X$ $\notin$ $\VAR(\atomc)$ (resp. $X$ $\notin$ $\VAR(\atoma')$). 

Inductively, we define $\atoma^{i+1}$ (resp. $\atomc^{i+1}$) to be the atom 
obtained from $\atoma^{i}$ (resp. $\atomc^{i}$) 
as follows: for each variable $X$ $\in$ $\VAR(\atomc)$ (resp. $X$ $\in$ $\VAR(\atoma')$), 
if each variables in positions
$\ARG(\atomc)\REST_{X}$ (resp. $\ARG(\atoma')\REST_{X}$) occurs as marked in 
$\atomc^{i}$ (resp. $\atoma^{i}$), then each occurrence of $X$ is marked in 
$\atoma^{i}$ (resp. $\atomc^{i}$) to obtain the new atom $\atomc^{i+1}$
(resp. $\atoma^{i+1}$).
Then naturally, we denote by $\atoma^\infty$ (resp. $\atomc^\infty$) 
as the fixpoint of the markup applications. Finally, we denote by 
$\MVAR(\atoma,\atomc,\atoma')$ as the set 
of all the marked variables mentioned \emph{only} in $\atoma^\infty$
under atoms $\atomc$ and $\atoma'$ as obtained through the method above.

Loosely speaking, in the aforementioned variable markup above, we can think of 
$\atoma$ as corresponding to some ``body atom" while $\atomc$ and 
$\atoma'$ as ``head atoms" that are reachable through the TGD extension 
$\tgds^+$ (see Definition \ref{tgd_extension}) as will respectively occur in some derivation track. 
Intuitively, the marked variables
represent element positions that may fail the ``sticky-join" property, i.e., disappear in the derivation track.
Intuitively, the sticky-join property insures decidability because only a finite number of elements can circulate
among the derivation tracks. 
As will be revealed in following Definition \ref{triangular_comp_def},
we further note that we only consider marked variables in terms of the
triple $\LA\atoma,\atomc,\atoma'\RA$ because we only consider them for ``recursive triangular-components."

\begin{definition}[\textbf{\textit{Recursive triangular-components}}]\label{triangular_comp_def}
	Let $\tgds$ be a set of TGDs and $\tgds^+$ its extension as defined in Definition \ref{tgd_extension}.
	Then a \emph{recursive triangular-component} (RTC) ${\cal T}$ is a tuple
	
	\vspace*{-0.4cm}
	
	\begin{align}
		&\hspace*{-0.1cm}\big(\LA B,H\RA,\LA\atoma,\atomb,\atomc\RA,\LA X,Z\RA,\atoma'\big),\label{triangular-component}
	\end{align}
	
	\vspace*{-0.1cm}	
	
	\noindent
	where: 
	\textbf{1.)} $\LA B,H\RA$ $\in$ $\tgds^+$;
	\textbf{2.)} $\{\atoma,\atomb\}$ $\subseteq$ $B$, $\atoma$ $\neq$ $\atomb$ and
	             $\atomc$ $\in$ $H$;
	\textbf{3.)} $\atoma'$ is an atom and there exists an assignment $\theta:$ $\VAR(\atoma)$ $\longrightarrow$ $\VAR(\atoma')$
	             such that $\atoma\theta$ $=$ $\atoma'$ and either one of the following holds:
	\begin{description}
	    \item[\rm (a)] $\atomc$ $=$ $\atoma'$, or
	    \item[\rm (b)] there exists $\LA B',H'\RA$ $\in$ $\tgds^+$ and function 
	             $\eta:$ $\VAR\big(B'\theta'\cup H'\theta')$ $\longrightarrow$ $\Gamma_{\cal C}\cup\Gamma_{\cal V}$,
	             where $\theta'$ is just a renaming substitution such that 
	             $\VAR(B'\theta'\cup H'\theta')$ $\cap$ $\VAR(B\cup H)$ $=$ $\emptyset$, and where
	             $\atomc$ $\in$ $B'(\eta\circ\theta')$ and $\atoma'$ $\in$ $H'(\eta\circ\theta')$;   
	\end{description}	
	\textbf{4.)} $X$ and $Z$ are two distinct variables where $\{X,Z\}$ $\subseteq$ $\widehat{\VAR}(B)$, and
	             $X$ $\in$ $\VAR(\atoma)$, $Z$ $\in$ $\VAR(\atomb)$, $\{X,Z\}$ $\subseteq$ $\VAR(\atomc)$ 
	             and $X$ $\in$ $\VAR(\atoma')$; and lastly, 
	\textbf{5.)} there exists a tuple of distinct atoms 
	             $\vatomd$ $=$ $\atomd_{1}\ldots\atomd_{m}$ $\subseteq$ $B$ such that:
    \begin{description}
	    \item[\rm (a)] $\atoma$ $=$ $\atomd_{1}$ and $\atomd_{m}$ $=$ $\atomb$, and for each $i$ $\in$ $\{1,\ldots,m-1\}$,
	                   there exists $Y_{i}$ $\in$ $\LINK(B,\,\atomd_{i},\,\atomd_{i+1})$;	     
	    %
	    \item[\rm (b)] for some $i$ $\in$ $\{1,\ldots,m-1\}$, there exists
					    $Y'$ $\in$ $\LINK(B,\,\atomd_{i},\,\atomd_{i+1})\mbox{\large $\setminus$}\{X,Z\}$ 
					    such that $Y'$ $\in$ $\VAR(\atoma')$ implies 
					    all occurrences of variables in positions $\ARG(\atoma')\REST_{Y'}$
					    in the atom $\atoma$ are in $\MVAR\big(\atoma,\vatomc,\atoma')$.
    \end{description}
\end{definition}

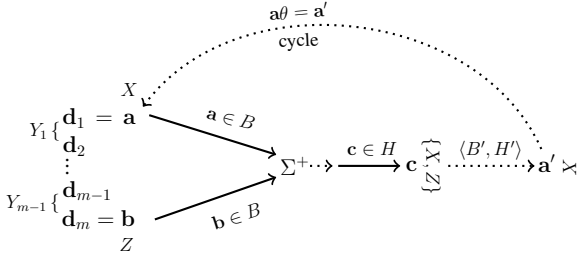
\begin{figure}
	\centering 
	\hspace*{-0.5cm}
	\begin{tikzpicture}[scale=0.75, every node/.style={scale=0.6}]
	   \begin{scope}[shift={(3cm,-5cm)}, fill opacity=1]
	   	   \node at (-5.64,0.85){\LARGE $\atomd_{1}\,=\,\atoma$};	  
	   	   \node at (-6.6,0.62){\Large $Y_{1}\,\{$};
	   	   \node at (-6.08,0.35){\LARGE $\atomd_{2}$};	  
	   	   \node at (-6.2,0.08){\LARGE $\vdots$};
	   	   \node at (-5.85,-0.45){\LARGE $\atomd_{m-1}$};	   	   
	   	   \node at (-6.81,-0.66){\Large $Y_{m-1}\,\{$};	   	   
	   	   \node at (-5.65,-0.95){\LARGE $\atomd_{m}=\atomb$};  
	   	   \draw[thick,->] (-4.75,0.90) to (-2.5,0.2);
	   	   \draw[thick,->] (-4.65,-0.95) to (-2.5,-0.2);
	   	   \node at (-2.17,0.0){\LARGE $\tgds^+$};
	   	   \draw[thick,dotted,->] (-1.9,0) to (-1.5,0.0);	   	   
	   	   \draw[thick,->] (-1.4,0.0) to (-0.3,0.0);
	   	   \draw[thick,dotted,->] (0.55,0.0) to (2.08,0.0);	   	   	
	   	   \node[rotate=0] at (1.3,0.28){\Large $\LA B',H'\RA$};	   	   
	   	   \draw[thick,dotted,->] (2.2,0.3) to [bend right=60] (-4.86,1.05);(1.7,0.05)
	   	   \node[rotate=-15] at (-3.3,0.85){\Large $\atoma\in B$};
	   	   \node[rotate=18] at (-3.2,-0.9){\Large $\atomb\in B$};	 
	   	   \node[rotate=0] at (-0.8,0.3){\Large $\atomc\in H$};
	   	   \node[rotate=-90] at (0.25,0.0){\Large $\{X,Z\}$};
	   	   \node[rotate=0] at (-0.1,0.0){\LARGE $\atomc$};
	   	   \node[rotate=-90] at (2.7,-0.03){\Large $X$};	   	   
	   	   \node[rotate=0] at (2.3,0.05){\LARGE $\atoma'$};	   	   	   	   	   	   
	   	   \node[rotate=0] at (-5.1,1.35){\Large $X$};	 
	   	   \node[rotate=0] at (-5.15,-1.4){\Large $Z$};	
	   	   \node[rotate=5] at (-2.1,2.75){\Large $\atoma\theta=\atoma'$};	   	   
	   	   \node[rotate=3] at (-2.1,2.2){\Large cycle};	   	   
	   \end{scope}
	\end{tikzpicture} 

	\vspace*{-0.3cm}
	
	\caption{ Recursive triangular-component (RTC).} 
	
	\vspace*{-0.4cm}
	        
    \label{triangular_comp_diagram} 
\end{figure}

%
Loosely speaking, a recursive triangular-component (RTC) ${\cal T}$ of the form (\ref{triangular-component})
(see Definition \ref{triangular_comp_def} and Figure \ref{triangular_comp_diagram}), 
can possibly enforce an infinite cycle of labeled nulls being ``pulled" together into a relation in
the chase derivation. We explain this by using again the TGDs $\tgds_2$ $=$ $\{\tgd_{11},\tgd^*_{12}\}$
and database $D_2$ of Example \ref{cyc_aag_examp}. Here, let us assume that $B$ $=$ $\BD(\tgd^*_{12})$
and $H$ $=$ $\HD(\tgd^*_{12})$ such that $\LA B,H\RA$ is the pair mentioned (\ref{triangular-component}).
Then with the body atoms $\SF{t}(X,Y)$, $\SF{u}(Y,Z)$ $\in$ $\BD(\tgd^*_{12})$ 
and head atom $\SF{t}(X,Z)$ $\in$ $\HD(\tgd^*_{12})$  
also standing for the atoms $\atoma$, $\atomb$ and $\atomc$ in (\ref{triangular-component}), respectively, 
then we can form the RTC:

\vspace*{-0.5cm}

\begin{align}
	&\hspace*{-0.2cm}\big(\big\LA B,H\RA,\big\LA\SF{t}(X,Y),\SF{u}(Y,Z),\SF{t}(X,Z)\big\RA,\LA X,Z\RA,\SF{t}(X,Z)\big).
	\label{examp_RTC}
\end{align}

\vspace*{-0.1cm}

We note here from Condition \textbf{3.)} of Definition \ref{triangular_comp_def} that the atom $\atomc'$
in (\ref{triangular-component}) is also the head atom ``$\SF{t}(X,Z)$", i.e., the choice (a) 
$\atomc$ $=$ $\atoma'$ of Condition \textbf{3.)} holds in this case.
For simplicity, we note that out example RTC in (\ref{examp_RTC}) retains the names of the variables ``$X$" and ``$Y$"
mentioned in (\ref{triangular-component}).
Loosely speaking, for two atoms $\SF{t}(\SF{n}_i,\SF{n}_j)$, $\SF{t}(\SF{n}_j,\SF{n}_k)$ 
$\in$ $\CHASE(D_2,\tgds_2)$, we have that rule $\tgd^*_{12}$ and its head atom ``$\SF{t}(X,Z)$"
would combine the two nulls ``$\SF{n}_i$" and ``$\SF{n}_j$" into a relation 
``$\SF{t}(\SF{n}_i,\SF{n}_j)$" in $\CHASE(D_2,\tgds_2)$. Since the variable ``$X$" is retained
in each RTC cycle via Condition \textbf{4.)} (see Figure \ref{triangular_comp_diagram}), this makes possible that
nulls held by ``$X$" in each cycle (in some substitution) to be pulled together into some other nulls held by ``$Z$"
as derived through the head atom ``$\SF{t}(X,Z)$". 

We further note that the connecting variable ``$Y$"
between the two body atoms ``$\SF{t}(X,Y)$" and ``$\SF{u}(Y,Z)$" corresponds to the variables $Y_i$
$\in$ $\LINK(B,\,\atomd_{i},\,\atomd_{i+1})$ of point (a) of Condition \textbf{5.)}, 
and for some $i$, some $Y'$ $\in$ $\LINK(B,\,\atomd_{i},\,\atomd_{i+1})\mbox{\large $\setminus$}\{X,Z\}$  
also appears as marked (i.e., $Y'$ $\in$ $\MVAR\big(\atoma,\atomc,\atoma')$) in 
point (b) of Condition \textbf{5.)} with respect to the atom $\atoma'$. 
Intuitively, we require in (b) of Condition \textbf{5.)} that \emph{some} of these variables $Y'$ occur as marked 
(w.r.t. $\atoma'$) so that labeled nulls of some link variables have a chance to disappear in the RTC cycle
for otherwise, they can only link and combine a bounded number of labeled nulls due to the 
\emph{sticky-join} property \cite{CaliGP12}.

\begin{example}\label{RTC_examp}
	Consider again the pair $p_{21}$ $=$ $\LA\{\SF{t}(X_1,V)$, $\SF{s}(V)$, $\SF{t}(V,Z_1)\}$, $\{\SF{t}(X_1,Z_1)\}\RA$ $\in$ $\tgds^2_3$ from Example \ref{tgd_extension_examp}. 
	Then with the pair $p_{21}$ standing for $\LA B,H\RA$ in (\ref{triangular-component}),
	the atoms ``$\SF{t}(X_1,V)$", ``$\SF{t}(V,Z_1)$", ``$\SF{t}(X_1,Z_1)$" and ``$\SF{t}(X_1,Z_1)$" for the atoms 
	$\atoma$, $\atomb$, $\atomc$ and $\atomc'$ in (\ref{triangular-component}), respectively, and
	variables $\LA X_1,Z_1\RA$ for the variables $\LA X,Z\RA$ in (\ref{triangular-component}), then we can get
	a corresponding RTC 
	${\cal T}_1$ $=$ 
	$\big(p_{21},\big\LA \SF{t}(X_1,V),\SF{t}(V,Z_1),\SF{t}(X_1,Z_1)\big\RA,\LA X_1,Z_1\RA,\SF{t}(X_1,Z_1)\big)$
	as illustrated in Figure \ref{triangular_comp_examp_diagram}.	
\end{example}
\begin{figure}
	\centering 
	\hspace*{-0.5cm}
	\begin{tikzpicture}[scale=0.75, every node/.style={scale=0.6}]
	\begin{scope}[shift={(3cm,-5cm)}, fill opacity=1]
	\node at (-5.64,0.85){\Large $\SF{t}(X_1,V)$};	  
	\node at (-6.78,0.5){\Large $V$};
	\node at (-5.85,0.0){\Large $\SF{s}(V)$};
	\draw[decorate,decoration={brace,mirror,raise=7pt}] (-6,0.9) to (-6,0.1);
	\draw[decorate,decoration={brace,raise=7pt}] (-6,-0.9) to (-6,0.0);		   	   
	\node at (-6.78,-0.5){\Large $V$};	   	   
	\node at (-5.65,-0.95){\Large $\SF{t}(V,Z_1)$};	   
	\draw[thick,->] (-4.95,0.90) to (-2.5,0.2);
	\draw[thick,->] (-4.95,-0.95) to (-2.5,-0.2);
	\node at (-2.17,0.0){\LARGE $\tgds^+_3$};
	\draw[thick,dotted,->] (-1.9,0) to (-1.5,0.0);	   	   
	\draw[thick,->] (-1.4,0.0) to (1.0,0.0);
	\draw[thick,dotted,->] (1.1,0.45) to [bend right=60] (-4.9,1.15);
	\node[rotate=-20] at (-3.4,0.8){\large $\SF{t}(X_1,V)\in B$};
	\node[rotate=18] at (-3.4,-0.9){\large $\SF{t}(V,Z_1)\in B$};	 
	\node[rotate=0] at (-0.2,0.25){\large $\SF{t}(X_1,Z_1)\in H$};		   	   
	\node[rotate=-90] at (2.65,0.0){\Large $\{X_1,Z_1\}$};
	\node[rotate=0] at (1.75,0.05){\Large $\SF{t}(X_1,Z_1)$};	   	   	   	   
	\node[rotate=0] at (-5.3,1.35){\Large $X_1$};	 
	\node[rotate=0] at (-5.3,-1.4){\Large $Z_1$};	
	\node[rotate=0] at (0.2,2.7){\Large $\SF{t}(X_1,V)\theta=\SF{t}(X_1,Z_1)$, 
		                          where $\theta$ $:=$ $\{X_1\mapsto X_1, V\mapsto Z_1\}$};	   	   
	\node[rotate=-3] at (-2.1,2.1){\Large cycle};	   	   
	\end{scope}
	\end{tikzpicture} 

	\vspace*{-0.2cm}
	
	\caption{RTC ${\cal T}_1$ of $\tgds_3$ with $\LA B,H\RA$ $=$ $p_{21}$ of Example \ref{tgd_extension_examp}.} 
	
	\vspace*{-0.4cm}
	
	\label{triangular_comp_examp_diagram} 
\end{figure}

\begin{definition}[\textbf{\textit{Triangularly-guarded TGDs}}]\label{TG_def}
    We say that a set of TGDs $\tgds$ is \emph{triangularly-guarded} (TG) iff for each
    RTC ${\cal T}$ of the form (\ref{triangular-component}) 
    (see Definition \ref{triangular_comp_def} and Figure \ref{triangular_comp_diagram}),
    we have that there exists some atom $\atomd$ $\in$ $B$ such that 
    $\{X,Z\}$ $\subseteq$ $\VAR(\atomd)$.
\end{definition}
\noindent
For convenience, we denote by $\TG$ as the class of all the triangularly-guarded TGDs.

\begin{example}
	Consider again the TGDs $\tgds_1$ in Example \ref{motive_examp} containing rules $\tgd_{11}$ and $\tgd_{12}$. 
	Then because there cannot be any derivation track that would combine the two variables ``$X$" and ``$Z$"
	of rule $\tgd_{12}$ into a single head atom in $\tgds^+_1$, then it follows that $\tgds_1$ cannot have any RTC. 
	Therefore, it trivially follows from Definition \ref{TG_def} that $\tgds_1$ is in the class $\TG$.
\end{example}

\section{Main Results and Concluding Remarks}

We now examine the important properties of this $\TG$ class of TGDs. 
In particular, we show that BCQ-Ans under the new class $\TG$ of TGDs is decidable.


\begin{definition}[\textbf{\textit{Interchangeable nulls}}]\label{int_nulls}
	Let $\vatoma$ $=$ $\atoma_1\ldots\atoma_l$ be a tuple of atoms where $\TERMS(\vatoma)$ $\subseteq$ $\Gamma_{\cal V}$, 
	$D$ be a database, $\tgds$ a set of TGDs 
	and $\SF{n}_{i}$, $\SF{n}_{j}$ $\in$ $\Gamma_{\SF{N}}$ \big($i,j$ $\in$ $\mathbb{N}$\big).
	Then we say that $\SF{n}_{i}$ and $\SF{n}_{j}$ are $\vatoma$-\emph{interchangeable} under 
	$\CHASE(D,\tgds)$ if for each \emph{connected} tuple of atoms 
	$\vatoma\theta$ $=$ $\theta(\atoma_1)\ldots\theta(\atoma_l)$ $\subseteq$ $\CHASE(D,\tgds)$, 
	where $\theta$ is a bijective (renaming) substitution,
	we have that $\{\SF{n}_{i},\SF{n}_{j}\}$ $\subseteq$ $\NULLS(\vatoma\theta)$
	implies 
	there exists an assignment $\theta':$ $\NULLS(\vatoma\theta)$ $\longrightarrow$ $\Gamma_\SF{N}$
	such that: (1) $\theta'(\SF{n}_{i})$ $=$ $\theta'(\SF{n}_{j})$; and 
	(2) $\vatoma(\theta'\circ\theta)$ $\subseteq$ $\CHASE(D,\tgds)$. 	
\end{definition}

Intuitively, with the tuple of atoms $\vatoma$ $=$ $\atoma_1\ldots\atoma_l$ as above, 
we have that $\SF{n}_i$ and $\SF{n}_j$ are ``$\vatoma$-interchangeable" 
under $\CHASE(D,\tgds)$ guarantees that if for some BCQ $Q$ $=$ $\exists\vect{X}\varphi(\vect{X})$ $\ra$ $q$ 
we have that $\CHASE(D,\tgds)$ $\models$ $Q$, then if $\varphi$ $=$ $\theta(\atoma_1)\wedge\ldots\wedge\theta(\atoma_l)$
for some renaming substitution $\theta$ (i.e., $\vatoma$ is the same ``type" as $\varphi$), 
then we have that simultaneously replacing all occurrences of $\SF{n}_j$ by $\SF{n}_i$ in $\CHASE(D,\tgds)$ 
would not affect the fact that $\CHASE(D,\tgds)$ $\models$ $Q$. 

Before we present the following Theorem \ref{bounded_nulls}, it is necessary to firstly introduce the notion \emph{level} in a chase that we define inductively as follows \cite{CaliGP12}: (1 ) for an atom $\atoma$ $\in$ $D$, we set $\LEVEL(\atoma)$ $=$ $0$; then inductively,
(2) for an atom $\atoma$ $\in$ $\CHASE(D,\tgds)$ obtained via some chase step $I_{k}\xrightarrow{\sigma,\,\eta}I_{k+1}$,
we set $\LEVEL(\atoma)$ $=$ $\MAX\big(\big\{\LEVEL(\atomb) \mid \atomb\in\BD(\tgd\eta)\}\big)$ $+$ $1$. Then finally,
for some given $k$ $\in$ $\mathbb{N}$, we set 
$\CHASE^k(D,\tgds)$ $=$ $\big\{\atoma\mid\atoma\in\CHASE(D,\tgds)\mbox{ and }\LEVEL(\atoma)\leq k\big\}$.
Intuitively, $\CHASE^k(D,\tgds)$ is the instance containing atoms that can be derived in a fewer or equal to 
$k$ chase steps. 

\begin{theorem}[\textbf{\textit{Bounded nulls}}]\label{bounded_nulls}
	Let $D$ be a database and $\tgds$ $\in$ $\TG$. Then for each
	tuple of atoms $\vatoma$, $\exists N$ $\in$ $\mathbb{N}$ 
	such that $\forall k$ $\in$ $\mathbb{N}$, we have that 
	$\SF{n}_j$ $\in$ $\big[\NULLS\big(\CHASE^{\,N+k}(D,\tgds)\big)$ $\mbox{\Large $\setminus$}$ $\NULLS\big(\CHASE^{\,N}(D,\tgds)\big)\big]$ implies 
	$\exists\SF{n}_i$ $\in$ $\NULLS\big(\CHASE^{\,N}(D,\tgds)\big)$ where $\SF{n}_i$ and $\SF{n}_j$ 
	are $\vatoma$-interchangeable under $\CHASE(D,\tgds)$.	
\end{theorem}
\begin{proofsketch}
	A contradiction can be derived by assuming that $\exists\vatoma$, $\forall N$ $\in$ $\mathbb{N}$, 
	$\exists k$ $\in$ $\mathbb{N}$, $\exists\SF{n}_j$ $\in$ $\Gamma_\SF{N}$, $\forall\SF{n}_i$ $\in$ $\Gamma_\SF{N}$, 
	where: $\SF{n}_j$ $\in$ $\big[\NULLS\big(\CHASE^{\,N+k}(D,\tgds)\big)$ {\Large $\setminus$} $\NULLS\big(\CHASE^{\,N}(D,\tgds)\big)\big]$, $\SF{n}_i$ $\in$ $\NULLS\big(\CHASE^{\,N}(D,\tgds)\big)$
	and $\SF{n}_i$ and $\SF{n}_j$ are \emph{not} $\vatoma$-interchangeable under $\CHASE(D,\tgds)$. Then the 
	fact that $\SF{n}_i$ and $\SF{n}_j$ are not $\vatoma$-interchangeable under $\CHASE(D,\tgds)$
	implies the existence of an infinite distinguishing relation among all those nulls $\SF{n}_i$ and $\SF{n}_j$.
	Therefore, it follows that there must exists some RTC ${\cal T}$ of the form (\ref{triangular-component}) 
	in $\tgds^+$ and where there are no body atom $\atomd$ that guards variables ``$X$" and ``$Z$", i.e., 
	$\{X,Z\}$ $\subseteq$ $\VAR(\atomd)$ (see Definition \ref{TG_def}).
\end{proofsketch}

\vspace*{-0.3cm}


%
\begin{theorem}[\textbf{\textit{Finite model property}}]\label{fm_thm}
	For database $D$, TGDs $\tgds$ $\in$ $\TG$ and BCQ $Q$, 
	$D$ $\cup$ $\tgds$ $\cup$ $\{\neg Q\}$ have the FM property.
\end{theorem}
\begin{proofsketch}
	Let $\vatoma$ $=$ $\atoma_1\ldots\atoma_l$ $=$ $\BD(x)$, for some
	$x$ $\in$ $\tgds\cup\{Q\}$. Then by Lemma \ref{bounded_nulls}, each of the null 
	$\SF{n}_j$ $\in$ $\big[\NULLS\big(\CHASE^{\,N+k}(D,\tgds)\big)$ $\mbox{\Large $\setminus$}$
	$\NULLS\big(\CHASE^{\,N}(D,\tgds)\big)\big]$ 
	is always $\vatoma$-interchangeable with some null $\SF{n}_i$ $\in$ $\NULLS\big(\CHASE^{\,N}(D,\tgds)\big)$.
	It then follows that $\CHASE(D,\tgds)$ can be represented by a finite number of nulls from which
	the finite model property follows.
\end{proofsketch}

\vspace*{-0.4cm}

\begin{theorem}[\textbf{\textit{Comparison with other syntactic classes}}]\label{contains_others}
    For each class ${\cal C}$ $\in$ $\{\SC{wa}$, 
    $\SC{w-guarded}$, $\SC{wsj}$, $\SC{g-guarded}$, $\SC{shy}$, $\SC{tame}$,  $\SC{wr}\}$,
    we have that ${\cal C}$ $\subsetneq$ $\TG$. 
\end{theorem}
\begin{proofsketch}
	A contradiction is derived by assuming that ${\cal C}$ $\in$ $\{\SC{wa}$, 
	$\SC{w-guarded}$, $\SC{wsj}$, $\SC{g-guarded}$, $\SC{shy}$, $\SC{tame}$,  $\SC{wr}\}$ but where
	${\cal C}$ $\notin$ $\TG$, since we have by Definition \ref{TG_def} that ${\cal C}$ $\notin$ $\TG$
	implies that there exists some RTC where the variables $X$ and $Z$ are not guarded by some 
	atom $\atomd$ $\in$ $B$.
\end{proofsketch}

\vspace*{-0.4cm}

\begin{theorem}[\textbf{\textit{Computational complexities}}]\label{membership_complexity}
	(1) Determining if $\tgds$ $\in$ $\TG$ is in $2$-\SC{ExpTime} (upper-bound) but is
	    \SC{Pspace}-hard (lower-bound); 
	(2) The BCQ-Ans combined complexity problem under the class $\TG$ is in $4$-\SC{ExpTime} (upper-bound)
	    but is $3$-\SC{ExpTime}-hard (lower-bound).		
\end{theorem}



In this paper, we have introduced a new class of TGDs called \emph{triangularly-guarded} TGDs 
($\TG$), for which BCQ-Ans is decidable as well as having the FM property (Theorems \ref{bounded_nulls} and 
\ref{fm_thm}). We further showed that $\TG$ strictly contains the current main syntactic classes: 
\SC{wa}, \SC{w-guarded}, \SC{wsj}, \SC{g-guarded},
\SC{shy}, \SC{tame} and \SC{wr} (Theorem \ref{contains_others}), which, to the best of our knowledge,
provides a unified representation of those aforementioned TGD classes.




\bibliographystyle{named}
\bibliography{TG_TGDs-IJCAI-2018}

\end{document}